\documentclass[10pt,twocolumn,letterpaper]{article}

\usepackage{iccv}
\usepackage{times}
\usepackage{epsfig}
\usepackage{graphicx}
\usepackage{amsmath}
\usepackage{amssymb}

\usepackage{multirow}
\usepackage{array}
\usepackage[linesnumbered,ruled]{algorithm2e}
\newcolumntype{L}[1]{>{\raggedright\let\newline\\\arraybackslash\hspace{0pt}}m{#1}}
\newcolumntype{C}[1]{>{\centering\let\newline\\\arraybackslash\hspace{0pt}}m{#1}}
\newcolumntype{R}[1]{>{\raggedleft\let\newline\\\arraybackslash\hspace{0pt}}m{#1}}

\usepackage[breaklinks=true,bookmarks=false]{hyperref}

\iccvfinalcopy % *** Uncomment this line for the final submission

\def\iccvPaperID{****} % *** Enter the ICCV Paper ID here

\begin{document}

%%%%%%%%% TITLE
\title{Generalized Coarse-to-Fine Visual Recognition with Progressive Training}

\author{Xutong Ren\textsuperscript{1}\thanks{This work was done when the first author was interning at Johns Hopkins University.},~ Lingxi Xie\textsuperscript{2,3}, Chen Wei\textsuperscript{1}, Siyuan Qiao\textsuperscript{2}\\Chi Su\textsuperscript{4}, Jiaying Liu\textsuperscript{1}, Qi Tian\textsuperscript{3}, Elliot K. Fishman\textsuperscript{5}, Alan L. Yuille\textsuperscript{2}\\
\textsuperscript{1}Peking University\quad\textsuperscript{2}Johns Hopkins University\quad\textsuperscript{3}Noah Ark's Lab, Huawei Inc.\\\textsuperscript{4}Kingsoft Cloud\quad\textsuperscript{5}Johns Hopkins Medical Institute\\
{\tt\small\{tonghelen,weichen582\}@pku.edu.cn}\quad{\tt\small\{198808xc,joe.siyuan.qiao\}@gmail.com}\quad{\tt\small suchi@kingsoft.com}\\{\tt\small liujiaying@pku.edu.cn}\quad{\tt\small tian.qi1@huawei.com}\quad{\tt\small efishman@jhmi.edu}\quad{\tt\small alan.yuille@jhu.edu}
}

\maketitle
%\thispagestyle{empty}

%%%%%%%%% ABSTRACT
\begin{abstract}
Computer vision is difficult, partly because the desired mathematical function connecting input and output data is often complex, fuzzy and thus hard to learn. Coarse-to-fine (C2F) learning is a promising direction, but it remains unclear how it is applied to a wide range of vision problems.

This paper presents a generalized C2F framework by making two technical contributions. First, we provide a unified way of C2F propagation, in which the coarse prediction (a class vector, a detected box, a segmentation mask, etc.) is encoded into a dense (pixel-level) matrix and concatenated to the original input, so that the fine model takes the same design of the coarse model but sees additional information. Second, we present a progressive training strategy which starts with feeding the ground-truth instead of the coarse output into the fine model, and gradually increases the fraction of coarse output, so that at the end of training the fine model is ready for testing. We also relate our approach to curriculum learning by showing that data difficulty keeps increasing during the training process. We apply our framework to three vision tasks including image classification, object localization and semantic segmentation, and demonstrate consistent accuracy gain compared to the baseline training strategy.
\end{abstract}

%%%%%%%%% BODY TEXT
\section{Introduction}
\label{Introduction}

Image recognition is a fundamental task of computer vision, which aims at understanding semantic contents from raw pixels. This is often difficult, because the underlying connection between low-level pixels and high-level semantics, {\em e.g.}, a mathematical function, is often complicated and fuzzy. For example, there exist a lot of elements in the data space which are either meaningless or ambiguous~\cite{nguyen2015deep}. In the deep learning era, researchers design neural networks as hierarchical and composite functions~\cite{krizhevsky2012imagenet,he2016deep}. However, the difficulty of training a network increases with its complexity~\cite{glorot2010understanding}. Despite some technical improvements designed to alleviate instability of training~\cite{nair2010rectified,srivastava2014dropout,ioffe2015batch}, a learned model can still suffer over-fitting, {\em i.e.}, it interprets a limited amount of training data in an improper way which does not generalize well to testing data.

This paper investigates visual recognition from the perspective of exploring a better training strategy. We focus on coarse-to-fine (C2F) learning, in which visual recognition takes two stages, with the first ({\em coarse}) stage producing rough prediction and the second ({\em fine}) one refining it towards higher accuracy. Mathematically, given a training pair $\left(\mathbf{x},\mathbf{y}^\star\right)$, the overall target function ${\mathbf{y}}={\mathbf{f}\!\left(\mathbf{x}\right)}$ is formulated into ${\mathbf{y}}={\mathbf{f}^\mathrm{F}\!\left(\mathbf{x},\mathbf{f}^\mathrm{C}\!\left(\mathbf{x}\right)\right)}$, with the superscripts $\mathrm{C}$ and $\mathrm{F}$ indicating `coarse' and `fine', respectively. It was believed that C2F learning strategy amortizes the difficulty of optimization and often leads to higher accuracy~\cite{li2013fixed,cao2015look}. However, it requires partitioning each recognition task into two stages, which is not always straightforward. For example, in image classification, it is unclear what the coarse stage can output as rough prediction, as well as how the rough prediction can be refined at the fine stage.

The key innovation of this paper lies in the first generalized C2F framework, to the best of our knowledge, that can be applied to a wide range of visual recognition tasks as well as a wide range of existing models, transforming them from direct optimization to C2F optimization. To achieve this goal, the essential difficulty lies in designing the fine model $\mathbf{f}^\mathrm{F}\!\left(\cdot\right)$, so that the input image $\mathbf{x}$ and coarse prediction $\mathbf{f}^\mathrm{C}\!\left(\mathbf{x}\right)$ are fused. We make {\bf our first contribution} as a unified way to achieve this goal, which encodes $\mathbf{f}^\mathrm{C}\!\left(\mathbf{x}\right)$ into a matrix $\mathbf{z}$ of the same spatial resolution as $\mathbf{x}$, and feeds the concatenated input data $\mathbf{x}\oplus\mathbf{z}$ into $\mathbf{f}^\mathrm{F}\!\left(\cdot\right)$. This is to say, $\mathbf{f}^\mathrm{F}\!\left(\cdot\right)$ can have the same architectural design as $\mathbf{f}^\mathrm{C}\!\left(\cdot\right)$ with merely the first (input) layer being slightly different. We instantiate our approach with three practical examples including image classification, object localization and semantic segmentation, and point out that many other vision problems can also be processed in a similar way.

In C2F learning, we point out two key issues that seem to conflict with each other. On the one hand, a joint optimization over the coarse and fine stages produces higher recognition accuracy~\cite{yu2018recurrent}. On the other hand, when the coarse model is not well optimized, the output $\mathbf{f}^\mathrm{C}\!\left(\mathbf{x}\right)$ can be noisy and impose a burden on optimizing the fine model. Thus, {\bf our second contribution} lies in a novel algorithm named progressive training~({\bf PT}) to alleviate this issue. The idea is to slightly modify the fine model as ${\mathbf{y}}={\mathbf{f}^\mathrm{F}\!\left(\mathbf{x},\tilde{\mathbf{y}}\right)}$, where $\tilde{\mathbf{y}}$ is sampled from a mixed distribution of $\mathbf{f}^\mathrm{C}\!\left(\mathbf{x}\right)$ and $\mathbf{y}^\star$ (ground-truth), and the probability of sampling $\mathbf{f}^\mathrm{C}\!\left(\mathbf{x}\right)$ is positively related to the elapsed iterations during training. In other words, $\mathbf{y}^\star$ is used to provide a warm start, but it is gradually replaced by $\mathbf{f}^\mathrm{C}\!\left(\mathbf{x}\right)$ and the training difficulty increases. At the end of training, the probability of sampling $\mathbf{y}^\star$ falls to $0$, so that the trained model can be applied to testing data. With mathematical analysis, we relate our idea to curriculum learning which aims at increasing data difficulty gradually during training.

As mentioned above, we apply our framework to three popular vision problems including image classification, object localization and semantic segmentation. Experimental results reveal consistent accuracy gain on top of the baseline, in particular when the amount of training data is limited, {\em i.e.}, the few-shot setting. Empirical analysis by comparing the training losses verifies our motivation, {\em i.e.}, such improvement comes from alleviating the risk of over-fitting.

The remainder of this paper is organized as follows. Section~\ref{RelatedWork} briefly reviews related work, and Section~\ref{Approach} presents our approach. After instantiating it on three visual recognition tasks in Section~\ref{Experiments}, we draw conclusions in Section~\ref{Conclusions}.

\section{Related Work}
\label{RelatedWork}

Deep learning~\cite{lecun2015deep} in particular deep convolutional neural networks have been dominating the field of computer vision. The fundamental idea is to build a hierarchical structure to learn complicated visual patterns from a large-scale database~\cite{deng2009imagenet}. As the number of network layers increases from tens~\cite{krizhevsky2012imagenet,simonyan2015very,szegedy2015going} to hundreds~\cite{he2016deep,huang2017densely}, the network's representation ability becomes stronger, but training these networks becomes more and more challenging. Various techniques have been proposed to improve numerical stability~\cite{nair2010rectified,ioffe2015batch} and over-fitting~\cite{srivastava2014dropout}, but the transferability from training data to testing data is still below satisfactory. It is pointed out that this issue is mainly caused by the overhigh complexity of deep networks, so that the limited amount of training data can be interpreted in an unexpected way~\cite{nguyen2015deep}. There exist two types of solutions, namely, coarse-to-fine learning and curriculum learning.

The idea of coarse-to-fine learning is based on the observation that a vision model can rethink its prediction to either predict fine-scaled information~\cite{moreels2005probabilistic} or amend errors~\cite{cao2015look}. Researchers designed several approaches for refining visual recognition in an iterative manner. These approaches can be explained using auto-context~\cite{tu2008auto} or formulated into a fixed-point model~\cite{li2013fixed}. Examples include the coarse-to-fine models for image classification~\cite{gangaputra2006design}, object detection~\cite{chen2016mitosis}, semantic segmentation~\cite{zhou2017fixed}, pose estimation~\cite{wei2016convolutional}, image captioning~\cite{gu2018stack}, {\em etc}. It is verified that joint training over coarse and fine stages boosts the performance~\cite{yu2018recurrent}, which raises an issue of the communication between coarse and fine stages in the training process -- we desire feeding coarse-stage output to fine-stage input, but when the coarse model has not been well optimized, this can lead to unstable performance in training. The method proposed in this paper can largely alleviate this issue.

Another idea, curriculum learning~\cite{bengio2009curriculum}, is aimed at gradually increasing the difficulty of training data, so that the training process becomes faster and/or more stable. This idea is first brought up by referring to how humans are taught to learn a concept and verified effective also for computer algorithms~\cite{khan2011humans}. It is later widely applied to a wide range of learning tasks, including visual recognition~\cite{sarafianos2017curriculum,tang2018attention} and generation~\cite{sharma2018improved}, natural language processing~\cite{li2016deep,ranzato2016sequence} and reinforcement learning~\cite{zaremba2015reinforcement,wu2017training}. Curriculum learning is theoretically verified a good choice in transfer learning~\cite{weinshall2018curriculum}, multi-task learning~\cite{pentina2015curriculum} and sequential learning~\cite{bengio2015scheduled} scenarios, and there have been discussions on the principle of designing curriculum towards better performance~\cite{zhou2018minimax}. A similar idea (gradually increasing training difficulty) is also adopted in online hard example mining (OHEM)~\cite{shrivastava2016training,chen2018sampleahead}, but OHEM often starts with a regular data distribution which is gradually adjusted towards difficult training data. The major drawback of curriculum learning lies in the requirement of evaluating the difficulty of training data, which is not easy in general. This paper provides a framework to bypass this problem.

\section{Our Approach}
\label{Approach}

This section describes our solution. We first present a unified framework of coarse-to-fine visual learning (our first contribution), and then propose a progressive algorithm towards effective training (our second contribution). In the final part, we draw the connection between this work and previous literature.

\subsection{A Generalized Coarse-to-Fine Framework}
\label{Approach:Coarse2Fine}

Let $\left(\mathbf{x},\mathbf{y}^\star\right)$ be a pair of training data, and the target function is ${\mathbf{y}}={\mathbf{f}\!\left(\mathbf{x};\boldsymbol{\theta}\right)}$. Throughout this paper, $\mathbf{f}\!\left(\cdot\right)$ is a deep neural network and $\boldsymbol{\theta}$ denotes the learnable parameters. A common framework of coarse-to-fine (C2F) learning is to decompose $\mathbf{f}\!\left(\mathbf{x};\boldsymbol{\theta}\right)$ into two stages, namely:
\begin{equation}
\label{Eqn:C2F}
{\mathbf{y}^\mathrm{F}}={\mathbf{f}^\mathrm{F}\!\left(\mathbf{x},\mathbf{y}^\mathrm{C};\boldsymbol{\theta}^\mathrm{F}\right)}={\mathbf{f}^\mathrm{F}\!\left(\mathbf{x},\mathbf{f}^\mathrm{C}\!\left(\mathbf{x};\boldsymbol{\theta}^\mathrm{C}\right);\boldsymbol{\theta}^\mathrm{F}\right)},
\end{equation}
where $\mathbf{f}^\mathrm{C}\!\left(\cdot\right)$ and $\mathbf{f}^\mathrm{F}\!\left(\cdot\right)$ are the coarse and fine models, respectively. In this way, we allow the coarse model $\mathbf{f}^\mathrm{C}\!\left(\cdot\right)$ to produce a rough prediction, and use the fine model $\mathbf{f}^\mathrm{F}\!\left(\cdot\right)$ to refine it. There exist some vision problems that are easy to be decomposed, {\em e.g.}, in semantic segmentation, one can apply a coarse stage to roughly locate the object followed by a fine stage to refine segmentation in a small region~\cite{zhou2017fixed,yu2018recurrent}. However, this is not always straightforward for an arbitrary vision task, {\em e.g.}, it is unclear how this formulation can work on image classification or object localization. The essential difficulty lies in combining $\mathbf{x}$ and $\mathbf{y}^\mathrm{C}$ in the fine model since they often have quite different dimensions.

We introduce a unified method which encodes $\mathbf{y}^\mathrm{C}$ into a matrix $\mathbf{z}$ of the same spatial resolution as $\mathbf{x}$, after which $\mathbf{x}$ and $\mathbf{z}$ are concatenated and fed into the fine model. This goal is often easy to achieve although the design varies from case to case. Here we instantiate it in three scenarios, each of which corresponds to a typical situation. In what follows, $W$ and $H$ denote the width and height of $\mathbf{x}$.

\vspace{-0.2cm}
\begin{itemize}
\item {\bf $\mathbf{y}^\mathrm{C}$ is a single value or class vector}, which happens in image classification. We duplicate $\mathbf{y}^\mathrm{C}$ by $W\times H$ times so as to fit the spatial resolution of input.
\vspace{-0.2cm}
\item {\bf $\mathbf{y}^\mathrm{C}$ corresponds to a set of bounding-boxes}, which happens in object localization. We construct a spatial map of a resolution of $W\times H$, and set the pixels inside the box to be $1$ while those outside to be $0$. This is easily generalized to multi-class detection, where we need to add one channel for each class.
\vspace{-0.2cm}
\item {\bf $\mathbf{y}^\mathrm{C}$ is a dense matrix}, which happens in semantic segmentation, edge detection, pose estimation, {\em etc}. We do not need any processing, {\em i.e.}, ${\mathbf{z}}={\mathbf{y}^\mathrm{C}}$.
\end{itemize}
\vspace{-0.2cm}
As a unified formulation, we denote ${\mathbf{z}}={\mathbf{g}\!\left(\mathbf{y}^\mathrm{C}\right)}$, so that Eqn~\eqref{Eqn:C2F} becomes ${\mathbf{y}^\mathrm{F}}={\mathbf{f}^\mathrm{F}\!\left(\mathbf{x}\oplus\mathbf{g}\!\left(\mathbf{f}^\mathrm{C}\!\left(\mathbf{x};\boldsymbol{\theta}^\mathrm{C}\right)\right);\boldsymbol{\theta}^\mathrm{F}\right)}$, where $\oplus$ denotes concatenation. The function $\mathbf{g}\!\left(\cdot\right)$, besides performing the above operation, 
has two convolutional layers, each of which is followed by a ReLU activation function~\cite{nair2010rectified}. The output of $\mathbf{g}\!\left(\cdot\right)$ has exactly the same dimensions as $\mathbf{x}$, which brings two-fold benefits.
First, it balances the amount of information provided by $\mathbf{x}$ and $\mathbf{g}\!\left(\mathbf{f}^\mathrm{C}\!\left(\mathbf{x};\boldsymbol{\theta}^\mathrm{C}\right)\right)$, otherwise the fine model is biased towards the latter part which often has a higher dimension. Second, the transformation prevents the fine model from directly taking the output and thus learning an identity mapping\footnote{This is more important with progressive training as shown in the following part, because at the start of training, we use the ground-truth $\mathbf{y}^\star$ instead of $\mathbf{y}^\mathrm{C}$ and thus it is easier to learn a dummy function $\mathbf{f}^\mathrm{F}\!\left(\cdot\right)$.}.

\subsection{Progressive Training}
\label{Approach:Optimization}

\newcommand{\figurewidth}{17cm}
\begin{figure*}[!t]
\centering
\includegraphics[width=\figurewidth]{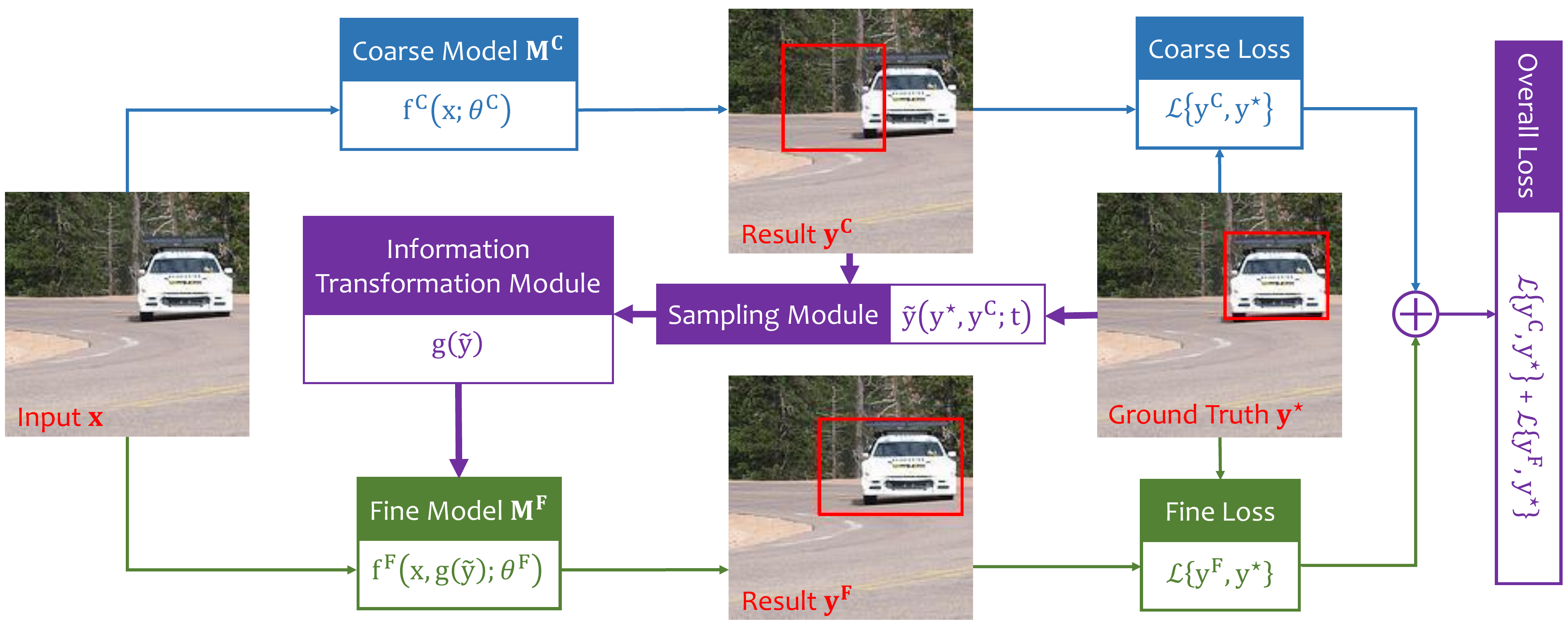}
\caption{Illustration of the proposed C2F framework with progressive training (best viewed in color), which is shown using an example of object localization. The coarse input $\mathbf{x}$ is fed into both the coarse and fine models, and the output of the coarse model is refined by the fine model towards higher accuracy. The sampling module $\tilde{\mathbf{y}}\!\left(\mathbf{y}^\star,\mathbf{y}^\mathrm{C};t\right)$ placed at the center is the key part of progressive training, which starts with ground-truth $\mathbf{y}^\star$ but gradually biases towards coarse prediction $\mathbf{y}^\mathrm{C}$ during the training process.}
\label{Fig:Framework}
\end{figure*}

Training the above C2F model involves optimizing a loss function, which is written in a generalized form:
\begin{equation}
\label{Eqn:OptimizationGoal}
\arg\min_{{\boldsymbol{\theta}^\mathrm{C},\boldsymbol{\theta}^\mathrm{F}}}\mathcal{L}\left\{\mathbf{f}^\mathrm{F}\!\left(\mathbf{x}\oplus\mathbf{g}\!\left(\mathbf{f}^\mathrm{C}\!\left(\mathbf{x};\boldsymbol{\theta}^\mathrm{C}\right)\right);\boldsymbol{\theta}^\mathrm{F}\right), \mathbf{y}^\star\right\}.
\end{equation}
Previous study~\cite{yu2018recurrent} points out the necessity of joint optimization which typically leads to higher accuracy. However, at the start of training, {\em i.e.}, when the coarse model is not well trained yet, ${\mathbf{y}^\mathrm{C}}={\mathbf{f}^\mathrm{C}\!\left(\mathbf{x};\boldsymbol{\theta}^\mathrm{C}\right)}$ often contains noisy information which raises extra challenges to optimizing the fine model. In practice, this often leads to instability in training, {\em e.g.}, the fine model fails to converge, or it mostly discards the coarse prediction so that the accuracy gain brought by the C2F framework is very much limited.

To deal with this problem, we make use of $\mathbf{y}^\star$ to compute $\mathbf{z}$ and feed it to the fine model at the beginning of training, but gradually replace $\mathbf{y}^\star$ with $\mathbf{y}^\mathrm{C}$ during the training process. To this end, we introduce a hyper-parameter ${t}\in{\left[0,1\right]}$, based on which a intermediate variable $\tilde{\mathbf{y}}$ is sampled from a mixed distribution of $\mathbf{y}^\mathrm{C}$ and $\mathbf{y}^\star$:
\begin{equation}
\label{Eqn:StochasticSampling}
{\tilde{\mathbf{y}}\!\left(\mathbf{y}^\star,\mathbf{y}^\mathrm{C};t\right)}=\left\{\begin{array}{ll}
\mathbf{y}^\star & \mathrm{if}\,{a\sim\mathcal{U}\!\left(0,1\right)}>{t} \\
\mathbf{y}^\mathrm{C} & \mathrm{otherwise}
\end{array}\right.,
\end{equation}
Here, $t$ is monotonically non-decreasing with respect to the number of training iterations, which is to say, the probability of feeding $\mathbf{y}^\star$ into the fine model is getting smaller with time\footnote{There are of course other options, {\em e.g.}, computing a weighted average like ${\tilde{\mathbf{y}}\!\left(\mathbf{y}^\star,\mathbf{y}^\mathrm{C};t\right)}={\left(1-t\right)\cdot\mathbf{y}^\star+t\cdot\mathbf{y}^\mathrm{C}}$. In experiments, we find that Eqn~\eqref{Eqn:StochasticSampling} shows the best overall performance, arguably because it ensures every sampled $\tilde{\mathbf{y}}$ is a real case (a weighted $\tilde{\mathbf{y}}$ may never happen).}. At the end of training, ${t}={0}$ and thus the fine model does not rely on the ground-truth at all, {\em i.e.}, the C2F model is ready for testing. This is named progressive training\footnote{The term of `progressive' is used in a few prior approaches~\cite{yang2015progressive,rusu2016progressive,liu2018progressive}, which has different motivations. Our goal is to gradually removing the use of ground-truth labels $\mathbf{y}^\star$ in the training process.}.

In summary, the overall loss function for training is:
\begin{equation}
\label{Eqn:LossFunction}
{\mathcal{L}}={\mathcal{L}\!\left\{\mathbf{f}^\mathrm{C}\!\left(\mathbf{x};\boldsymbol{\theta}^\mathrm{C}\right),\mathbf{y}^\star\right\}+\mathcal{L}\!\left\{\mathbf{f}^\mathrm{F}\!\left(\mathbf{x}\oplus\mathbf{z};\boldsymbol{\theta}^\mathrm{F}\right),\mathbf{y}^\star\right\}}.
\end{equation}
\iffalse
\begin{eqnarray}
\nonumber
{\mathcal{L}}={\mathcal{L}\!\left\{\mathbf{f}^\mathrm{C}\!\left(\mathbf{x};\boldsymbol{\theta}^\mathrm{C}\right),\mathbf{y}^\star\right\}+\mathcal{L}\!\left\{\mathbf{f}^\mathrm{F}\!\left(\mathbf{x}\oplus\mathbf{z};\boldsymbol{\theta}^\mathrm{F}\right),\mathbf{y}^\star\right\}}\\
\label{Eqn:LossFunction}
\quad={\left|\mathbf{f}^\mathrm{C}\!\left(\mathbf{x};\boldsymbol{\theta}^\mathrm{C}\right)-\mathbf{y}^\star\right|^2+\left|\mathbf{f}^\mathrm{F}\!\left(\mathbf{x}\oplus\mathbf{g}\!\left(\tilde{\mathbf{y}}\right);\boldsymbol{\theta}^\mathrm{F}\right)-\mathbf{y}^\star\right|^2}.
\end{eqnarray}
\fi
When $\mathbf{y}^\mathrm{C}$ is chosen to be $\tilde{\mathbf{y}}$, the gradient of the second term involves both $\boldsymbol{\theta}^\mathrm{C}$ and $\boldsymbol{\theta}^\mathrm{F}$ and the coarse and fine models are optimized jointly. We add a coarse loss term so that $\boldsymbol{\theta}^\mathrm{C}$ is better optimized~\cite{szegedy2015going,lee2015deeply} and the stability of training is higher~\cite{yu2018recurrent}. The overall framework is shown in Figure~\ref{Fig:Framework}.

From another perspective, our approach provides a tradeoff between training stability and generalization ability. When $\mathbf{y}^\star$ is fed into the fine model, training becomes easier as accurate cues are received, but the trained model is more difficult to transfer due to the unavailability of $\mathbf{y}^\star$ during testing. We alleviate this issue via a gradual transition from $\mathbf{y}^\star$ to $\mathbf{y}^\mathrm{C}$. As we shall see in experiments, this overcomes over-fitting and leads to higher recognition accuracy.

\subsection{Summary and Relationship to Prior Works}
\label{Approach:Relationship}

In summary, we make two contributions, with the first one being an encoding scheme for generalized C2F formulation, and the second one being a progressive training strategy that improves both stability and accuracy of training. {\bf To the best of our knowledge, this is the first C2F framework that can be freely applied to a wide range of visual recognition tasks.}

It is necessary to discuss the relationship between our approach and curriculum learning~\cite{bengio2009curriculum} which aims at gradually increasing the difficulty of data during the training process. We show that progressive training, though not being a strict curriculum process (according to the definitions in~\cite{bengio2009curriculum}), has quite similar properties. To reveal this, let ${\mathbf{u}}={\left(\mathbf{x},\tilde{\mathbf{y}}\right)}$ be an input of $\mathbb{M}^\mathrm{F}$, which can be sampled from the {\em coarse-prediction distribution} $\mathcal{P}_{\mathbb{M}_t^\mathrm{C}}$:
\begin{equation}
{P_{\mathbb{M}_t^\mathrm{C}}\!\left(\mathbf{u};\boldsymbol{\theta}_t^\mathrm{C}\right)}={P\!\left(\mathbf{x}\right)\cdot\mathcal{N}\!\left(\tilde{\mathbf{y}}\mid\mathbf{f}^\mathrm{C}\!\left(\mathbf{x};\boldsymbol{\theta}_t^\mathrm{C}\right),\sigma_t\mathbf{I}\right)},
\end{equation}
where $P\!\left(\mathbf{x}\right)$ is determined by the training set and $\mathcal{N}\!\left(\cdot\right)$ is an isotropic Gaussian distribution, which degenerates to the Kronecker $\delta$-function when its variance ${\sigma}\rightarrow{0}$. Similarly, we define the {\em ground-truth distribution} $\mathcal{P}_\mathrm{GT}$:
\begin{equation}
{P_\mathrm{GT}\!\left(\mathbf{u}\right)}={P\!\left(\mathbf{x}\right)\cdot\mathcal{N}\!\left(\tilde{\mathbf{y}}\mid\mathbf{y}^\star,\sigma_t\mathbf{I}\right)}.
\end{equation}
In the training process, $\mathcal{P}_{\mathbb{M}_t^\mathrm{C}}$ changes with $\boldsymbol{\theta}_t^\mathrm{C}$ while $\mathcal{P}_\mathrm{GT}$ remains constant. Let $\mathcal{P}_t$ denote the distribution of $\mathbf{u}$ at time $t$. We make a simple assumption that the difference between $\mathcal{P}_\mathrm{GT}$ and $\mathcal{P}_{\mathbb{M}_t^\mathrm{C}}$ is relatively large\footnote{This can be explained as (i) at the early training stage, coarse prediction is often less accurate, {\em i.e.}, $\mathbf{y}^\star$ is often far away from $\mathbf{f}^\mathrm{C}\!\left(\mathbf{x};\boldsymbol{\theta}_t^\mathrm{C}\right)$, while (ii) at the late training stage, coarse prediction becomes more accurate but also more deterministic, {\em i.e.}, $\sigma_t$ becomes very small.}, so that we can approximate the Shannon entropy of $\mathcal{P}_t$ as:
\begin{eqnarray}
{P_t\!\left(\mathbf{u}\right)}={\left(1-t\right)\cdot P_\mathrm{GT}\!\left(\mathbf{u}\right)+t\cdot P_{\mathbb{M}_t^\mathrm{C}}\!\left(\mathbf{u};\boldsymbol{\theta}_t^\mathrm{C}\right)},\\
{\mathbb{H}\!\left[\mathcal{P}_t\right]}\approx{\left(1-t\right)\cdot\mathbb{H}\!\left[\mathcal{P}_\mathrm{GT}\right]+t\cdot\mathbb{H}\!\left[\mathcal{P}_{\mathbb{M}_t^\mathrm{C}}\right]+\mathbb{H}\!\left(t\right)},
\end{eqnarray}
where ${\mathbb{H}\!\left(t\right)}={-t\ln t-\left(1-t\right)\ln\!\left(1-t\right)}$, which has an upper bound of $\ln2$. On the other hand, $\mathbb{H}\!\left[\mathcal{P}_\mathrm{GT}\right]$ is smaller than $\mathbb{H}\!\left[\mathcal{P}_{\mathbb{M}_t^\mathrm{C}}\right]$ by a margin, because $\mathcal{P}_\mathrm{GT}$ is constant. Thus, during the training process, $\mathbb{H}\!\left[\mathcal{P}_t\right]$ is mostly increasing, which implies that training difficulty becomes larger.

Other closely related works to our approach include~\cite{ranzato2016sequence} and~\cite{yu2018recurrent}. \cite{ranzato2016sequence} considers a sequence learning task in which each cell takes the output of the previous cell as input. In each training epoch, the first part of training data are provided by the ground-truth while the second provided by the prediction, and the fraction was controlled by the elapsed training time $t$. Differently, progressive training allows the data distribution to be changed more smoothly and thus improves training stability. \cite{yu2018recurrent} proposes a coarse-to-fine framework for semantic segmentation, and uses a weaker version of curriculum learning in which the distribution was changed from $\mathcal{P}_\mathrm{GT}$ to $\mathcal{P}_{\mathbb{M}_t^\mathrm{C}}$ all at once. This sudden change may cause the model to fail to convergence. With progressive training, the distribution is gradually changed during training, leading to consistent convergence and accuracy gain in experiments (see Section~\ref{Experiments:Segmentation}).

\section{Applications}
\label{Experiments}

In this section, we apply the designed framework to three popular vision tasks, including object localization (Section~\ref{Experiments:Localization}), image classification (Section~\ref{Experiments:Classification}) and semantic segmentation (Section~\ref{Experiments:Segmentation}), each of which corresponds to a type of encoding $\mathbf{y}^\mathrm{C}$ as discussed in Section~\ref{Approach:Coarse2Fine}. We also show that the improvement brought by our approach transfers smoothly from object localization to object parsing.

\subsection{Object Localization}
\label{Experiments:Localization}

\subsubsection{Settings}
\label{Experiments:Localization:Settings}

In the first part, we apply our framework to object localization, which differs from object detection in that we do not need to predict the object class in both training and testing -- for each input image, the desired output is a bounding box that indicates the object. While being less specific, this system can assist a wide range of vision tasks including object detection~\cite{qiao2017scalenet} and object parsing, {\em i.e.}, detecting the semantic parts of an object~\cite{zhang2018deepvoting}. Here, we assume that only one object exists in each image, but as shown in~\cite{zhang2018deepvoting}, this assumption can be easily taken out by applying simple techniques in the testing process.

We take ScaleNet~\cite{qiao2017scalenet} as our baseline and collect data from the ILSVRC2012 dataset~\cite{russakovsky2015imagenet}, in which $21$ categories with the superclass of {\em vehicle} are chosen, since the original method provides reasonable prediction on rigid objects. We only choose those training and testing images with exactly one bounding-box annotated\footnote{In the ILSVRC2012 dataset, about half of training images are not annotated with bounding-boxes, but all testing images are annotated with bounding-boxes.}, and ignore those with more than one objects annotated to avoid confusion. In total, there are around $10\rm{,}000$ training and $1\rm{,}000$ testing images.

Following~\cite{qiao2017scalenet}, the entire image is rescaled into $192\times192$ with its aspect ratio preserved. Empty stripes are added if necessary. Then it is fed into a $50$-layer deep residual network~\cite{he2016deep} (only the middle part with $39$ layers is actually used). The output consists of four floating point numbers, indicating the coordinate $\left(x,y\right)$ of the central pixel, the width $W$ and the height $H$ of the bounding box, respectively. These numbers are individually compared with the ground-truth using the log-scale $\ell_1$-norm and summed up to the final loss. At the testing stage, we compute two evaluation metrics, namely, the IOU between the ground-truth and predicted bounding boxes, and an accuracy indicating whether the IOU is at least $0.5$.

\subsubsection{Different Learning Options}
\label{Experiments:Localization:Options}

\newcommand{\colwidthA}{0.80cm}
\newcommand{\colwidthB}{0.50cm}
\newcommand{\colwidthC}{1.00cm}
\begin{table}[!btp]
\centering{
\begin{tabular}{|L{\colwidthA}||C{\colwidthB}|C{\colwidthB}|C{\colwidthC}||C{\colwidthC}|C{\colwidthC}|}
\hline
{}        & $t_0$ & $E$  & STG &   IOU  &   Acc. \\
\hline\hline
{\bf BL}  & $-$   & $-$  & N/A &           $74.85$ &           $85.74$ \\
\hline
{\bf IND} & $-$   & $-$  & F   &           $75.18$ &           $86.43$ \\
\hline
\multirow{2}{*}{{\bf JNT}} & \multirow{2}{*}{$-$} & \multirow{2}{*}{$-$}
             & C   &           $73.97$ &           $84.38$ \\
\cline{4-6}
{} & {} & {} & F   &           $75.60$ &           $86.43$ \\
\hline\hline
\multirow{2}{*}{{\bf PT}}  & \multirow{2}{*}{$0.0$} & \multirow{2}{*}{$30$}
             & C   &           $74.77$ &           $84.96$ \\
\cline{4-6}
{} & {} & {} & F   &           $75.35$ &           $85.94$ \\
\hline
\multirow{2}{*}{{\bf PT}}  & \multirow{2}{*}{$0.5$} & \multirow{2}{*}{$30$}
             & C   &           $75.07$ &           $85.64$ \\
\cline{4-6}
{} & {} & {} & F   &  $\mathbf{77.38}$ &  $\mathbf{87.60}$ \\
\hline
\multirow{2}{*}{{\bf PT}}  & \multirow{2}{*}{$0.5$} & \multirow{2}{*}{$50$}
             & C   &           $74.31$ &           $85.94$ \\
\cline{4-6}
{} & {} & {} & F   &           $75.04$ &           $86.13$ \\
\hline
\end{tabular}}\vspace{2mm}
\caption{Object localization accuracy ($\%$) on the ILSVRC2012 {\em vehicle} superclass using different learning approaches and different options of our approach. $t_0$ and $E$ indicate the value of $t$ at the start of training and the number of epochs when $t$ reaches $1.0$. STG indicates the stage (coarse or fine). The coarse stage of {\bf IND} is {\bf BL}. There are $50$ training epochs in total. For detailed descriptions of these evaluation metrics, see Section~\ref{Experiments:Localization:Settings}.}
\label{Tab:ObjectLocalization}
\end{table}

\renewcommand{\figurewidth}{8cm}
\begin{figure}
\centering
\includegraphics[width=\figurewidth]{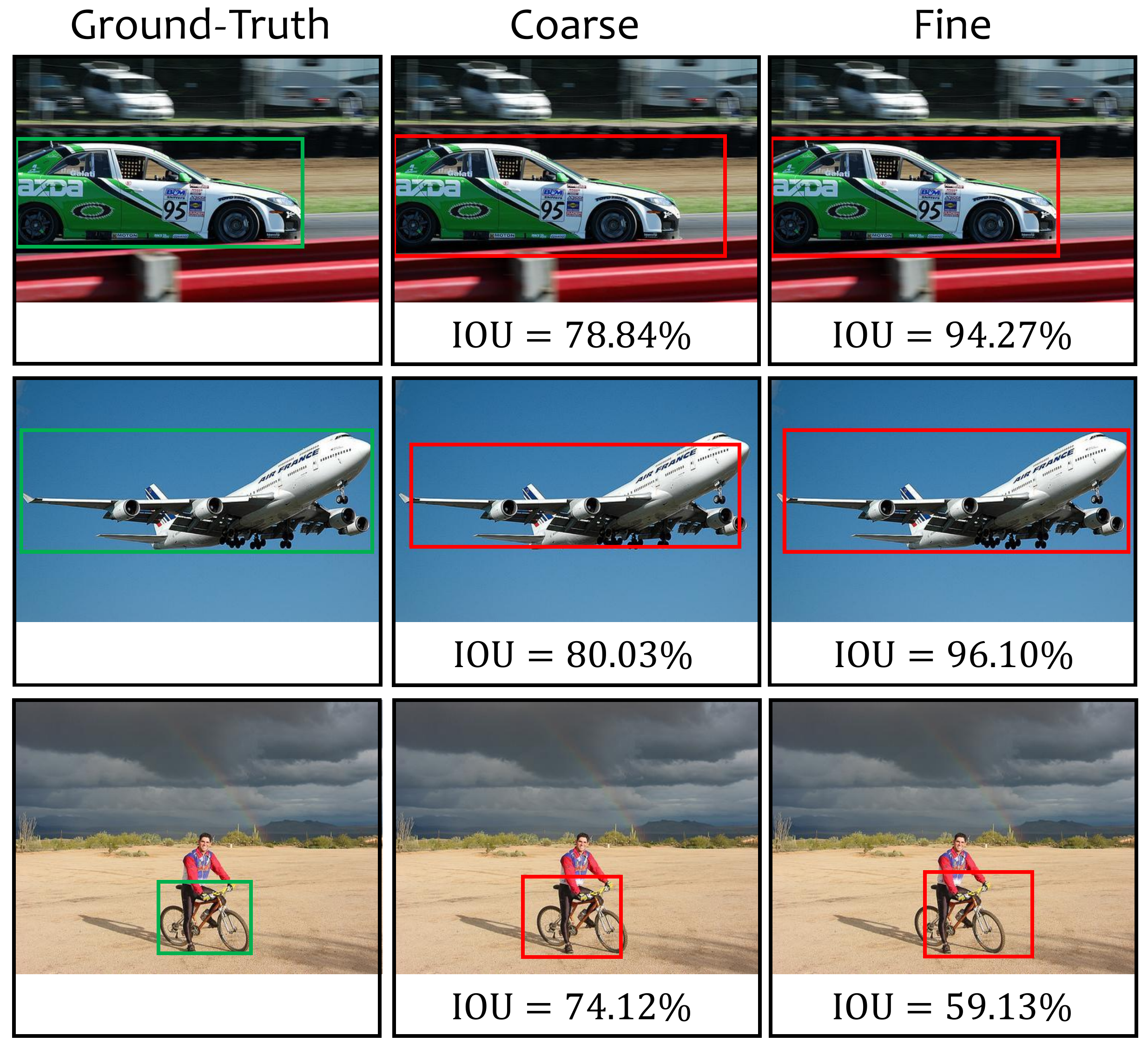}
\caption{Two successful and one failure cases of our approach in object localization (best viewed in color). The first and second rows show the successful cases. The bottom row shows a failure case. In each row, the green frame indicates the ground-truth, and the red frame indicates the prediction.}
\label{Fig:Localization}
\end{figure}

To study this task in a coarse-to-fine manner, we first construct a weighted map using the predicted $x$, $y$, $W$ and $H$ from the coarse stage. The values within the bounding box is set to be $1$ and those outside set to be $0$. This map is then passed through two convolutional layers and appended to $\mathbf{x}$. Although this box only provides a limited amount of information, we shall see the improvement it brings to object localization.

We compare progressive training ({\bf PT}) with other three training strategies. The baseline ({\bf BL}) simply trains one single network, {\em a.k.a.}, the coarse model. The individual training ({\bf IND}) and joint training ({\bf JNT}) methods train the coarse and fine model simultaneously, but in individual and joint manners, respectively. Here, by joint training we mean to provide $\mathbf{y}^\mathrm{C}$ to $\mathbb{M}^\mathrm{F}$ from the beginning of training, {\em i.e.}, ${t}\equiv{1}$, while individual training means to provide $\mathbf{y}^\star$ during the entire training stage, {\em i.e.}, ${t}\equiv{0}$. Moreover, we study different options of {\bf PT} defined by $t_0$ (the $t$ value at the start of training) and $E$ (the number of epochs when $t$ reaches $1$), and we assume that $t$ always grows linearly with training time.

Results are summarized in Table~\ref{Tab:ObjectLocalization}. Two interesting phenomena are observed. First, starting training with a non-zero $t$ often improves performance, since when ${t}={0}$, the extra information is too strong so that the fine model can be severely biased towards such ``cheating'' information and thus learns a weaker connection between image data and output labels. In addition, it is always better for the model to be trained on ${t}\equiv{1}$ (the same setting as in testing) for several epochs, so that the model can adjust to this scenario. Examples showing how our approach works, including a failure case of localization, are in Figure~\ref{Fig:Localization}.

In Figure~\ref{Fig:Curves}, we compare the learning curves of {\bf IND}, {\bf JNT} and the best {\bf PT} (${t_0}={0.5}$, ${E}={30}$). We can see that the fine phase of {\bf IND} achieves a very low training error by heavily over-fitting training data, in particular, with the ``cheating'' information from $\mathbf{y}^\star$. The loss of {\bf JNT} is much higher at the beginning, because the fine stage is confused by the coarse stage. As training continues, the loss term becomes smaller because it starts fitting the coarse prediction. This does not bring benefits, because the potential errors in coarse prediction are not fixed. {\bf PT} alleviates this issue by starting with a relatively easy task in which parts of data are assisted by the ground-truth, and gradually moving onto the real distribution, during which the ground-truth data are still provided to prevent the model from being impacted by inaccurate coarse predictions.

\renewcommand{\figurewidth}{8cm}
\begin{figure}
\centering
\includegraphics[width=\figurewidth]{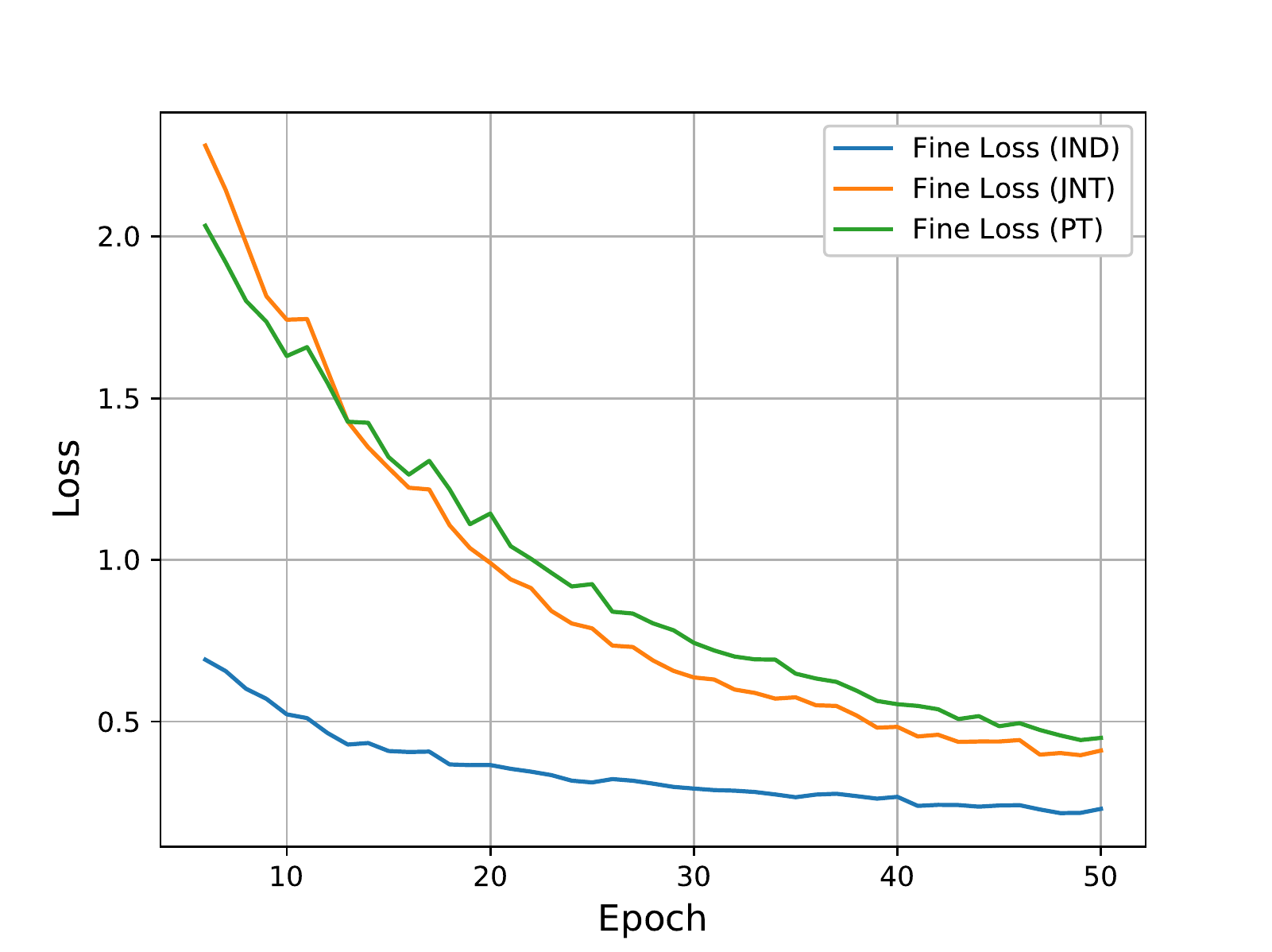}
\caption{Learning curves of {\bf IND}, {\bf JNT} and {\bf PT} with ${t_0}={0.5}$ and ${E}={30}$ (best viewed in color).}
\label{Fig:Curves}
\end{figure}

\subsubsection{Application to Object Parsing}
\label{Experiments:Localization:Application}

Finally, we apply object localization results to object parsing, {\em i.e.}, detecting the so-called semantic parts in objects. Here, each semantic part refers to a verbally describable pattern in the object, {\em e.g.}, the {\em wheels} of a {\em car} or the {\em pedal} of a {\em bike}. We use DeepVoting~\cite{zhang2018deepvoting}, which requires all training objects to have a fixed scale. Thus, accurate object localization (either scale and location) helps a lot.

We use the VehicleSemanticParts (VSP) dataset introduced in~\cite{zhang2018deepvoting}, which is created from the {\em vehicle} images in Pascal3D+~\cite{xiang2014beyond}. There are six types of vehicles, namely, {\em airplane}, {\em bike}, {\em bus}, {\em car}, {\em motorbike} and {\em train}. There are different numbers of semantic parts annotated for each class, and we directly use the trained models for these six classes, {\em i.e.}, DeepVoting itself is not modified, and we only change the scale prediction module which aims at providing a better input for DeepVoting. At the testing stage, we also add random occlusion by extracting pixel-wise masks from irrelevant objects ({\em e.g.}, {\em cat} or {\em dog}) in the PascalVOC 2007 dataset~\cite{everingham2010pascal} and placing them on the input images. By controlling the number of occluders and the fraction of occlusion, we construct four levels of difficulties denoted by {\bf L0}, {\bf L1}, {\bf L2} and {\bf L3}, with {\bf L0} indicating no occlusion, and {\bf L3} the heaviest occlusion.

\renewcommand{\colwidthA}{0.54cm}
\renewcommand{\colwidthB}{0.76cm}
\begin{table}[!btp]
\centering{
\setlength{\tabcolsep}{0.08cm}
\begin{tabular}{|L{\colwidthA}||C{\colwidthB}|C{\colwidthB}||C{\colwidthB}|C{\colwidthB}||C{\colwidthB}|C{\colwidthB}||C{\colwidthB}|C{\colwidthB}|}
\hline
\multirow{2}{*}{SC}
          & \multicolumn{2}{c||}{{\bf L0}} & \multicolumn{2}{c||}{{\bf L1}} & \multicolumn{2}{c||}{{\bf L2}} &  \multicolumn{2}{c|}{{\bf L3}} \\
\cline{2-9}
{}        &   {\bf BL} &          {\bf PT} &   {\bf BL} &          {\bf PT} &   {\bf BL} &          {\bf PT} &   {\bf BL} &          {\bf PT} \\
\hline\hline
{\em ai.} &     $62.4$ &   $\mathbf{75.4}$ &     $46.1$ &   $\mathbf{60.9}$ &     $46.6$ &   $\mathbf{59.3}$ &     $44.6$ &   $\mathbf{55.8}$ \\
\hline
{\em bi.} &     $66.8$ &   $\mathbf{77.2}$ &     $64.0$ &   $\mathbf{74.8}$ &     $52.3$ &   $\mathbf{68.5}$ &     $46.7$ &   $\mathbf{61.1}$ \\
\hline
{\em bu.} &     $81.5$ &   $\mathbf{86.9}$ &     $57.2$ &   $\mathbf{80.9}$ &     $54.0$ &   $\mathbf{68.8}$ &     $47.8$ &   $\mathbf{61.5}$ \\
\hline
{\em ca.} &     $89.7$ &   $\mathbf{96.0}$ &     $67.2$ &   $\mathbf{71.8}$ &     $58.2$ &   $\mathbf{61.0}$ &     $\mathbf{49.6}$ &   $47.0$ \\
\hline
{\em mo.} &     $59.9$ &   $\mathbf{73.4}$ &     $52.8$ &   $\mathbf{63.1}$ &     $42.8$ &   $\mathbf{55.9}$ &     $38.5$ &   $\mathbf{53.1}$ \\
\hline
{\em tr.} &     $55.1$ &   $\mathbf{74.0}$ &     $47.4$ &   $\mathbf{62.9}$ &     $42.0$ &   $\mathbf{59.2}$ &     $37.5$ &   $\mathbf{52.3}$ \\
\hline\hline
{\bf avg} &     $69.2$ &   $\mathbf{80.5}$ &     $55.8$ &   $\mathbf{69.1}$ &     $49.3$ &   $\mathbf{62.1}$ &     $44.1$ &   $\mathbf{55.1}$ \\
\hline
\hline
\multirow{2}{*}{SP}
          & \multicolumn{2}{c||}{{\bf L0}} & \multicolumn{2}{c||}{{\bf L1}} & \multicolumn{2}{c||}{{\bf L2}} &  \multicolumn{2}{c|}{{\bf L3}} \\
\cline{2-9}
{}        &   {\bf BL} &          {\bf PT} &   {\bf BL} &          {\bf PT} &   {\bf BL} &          {\bf PT} &   {\bf BL} &          {\bf PT} \\
\hline\hline
{\em ai.} &     $60.3$ &   $\mathbf{61.3}$ &     $40.6$ &   $\mathbf{42.6}$ &     $32.3$ &   $\mathbf{34.8}$ &     $25.4$ &   $\mathbf{27.4}$ \\
\hline
{\em bi.} &     $90.8$ &   $\mathbf{92.3}$ &     $85.2$ &   $\mathbf{88.4}$ &     $79.6$ &   $\mathbf{81.6}$ &     $62.5$ &   $\mathbf{67.8}$ \\
\hline
{\em bu.} &     $81.3$ &   $\mathbf{82.1}$ &     $65.8$ &   $\mathbf{65.9}$ &     $\mathbf{54.6}$ &   $54.3$ &     $\mathbf{40.5}$ &   $39.9$ \\
\hline
{\em ca.} &     $80.6$ &   $\mathbf{81.1}$ &     $57.3$ &   $\mathbf{57.6}$ &     $\mathbf{41.7}$ &   $41.3$ &     $\mathbf{29.4}$ &   $28.0$ \\
\hline
{\em mo.} &     $\mathbf{69.7}$ &   $69.3$ &     $55.5$ &   $\mathbf{56.5}$ &     $43.4$ &   $\mathbf{46.5}$ &     $31.2$ &   $\mathbf{34.3}$ \\
\hline
{\em tr.} &     $61.2$ &   $\mathbf{65.3}$ &     $43.7$ &   $\mathbf{45.8}$ &     $29.8$ &   $\mathbf{31.6}$ &     $\mathbf{22.2}$ &   $22.1$ \\
\hline\hline
{\bf avg} &     $74.0$ &   $\mathbf{75.2}$ &     $58.0$ &   $\mathbf{59.4}$ &     $46.9$ &   $\mathbf{48.4}$ &     $35.2$ &   $\mathbf{36.6}$ \\
\hline
\end{tabular}}\vspace{2mm}
\caption{Scale (SC) prediction and semantic part (SP) detection accuracy ($\%$) on the VSP dataset, measured by a threshold of $10\%$ relative difference and mean average precision (mAP), respectively. {\bf L0}, {\bf L1}, {\bf L2} and {\bf L3} indicate different occlusion levels.}
\label{Tab:SemanticPartDetection}
\end{table}

We train two scale prediction models {\bf BL} and {\bf PT} (${t_0}={0.5}$, ${E}={30}$) on the training set of VSP, in which each image provides a bounding box for the only {\em vehicle} in it. On the testing set, we compute both scale prediction accuracy, measured by whether it differs from the ground-truth by more than $10\%$, which follows the original work~\cite{zhang2018deepvoting} and semantic part detection accuracy, measured by mAP. Results are summarized in Table~\ref{Tab:SemanticPartDetection}. We can see that, our approach generalizes from ILSVRC2012 to Pascal3D+ well for scale prediction, and the more accurate scale prediction indeed helps object parsing, {\em i.e.}, the improvement of mAP, averaged over six classes, exceeds $1\%$ at all occlusion levels. This demonstrates a wide application of our approach.

\subsection{Image Classification}
\label{Experiments:Classification}

We next investigate a fundamental problem, image classification, from two perspectives, namely, regular classification and few-shot classification. The coarse-to-fine algorithm works similarly as that in object localization. We copy the classification results, a $C$-dimensional vector with $C$ being the number of classes, by $W\times H$ times, pass it through two convolutional layers, and append it to the original input. We follow the optimal strategy learned from object localization, which starts with ${t_0}={0.5}$ and arrives at ${t_0}={1}$ in the midst of training.

\subsubsection{Regular Classification}
\label{Experiments:Classification:Regular}

Regular experiments are performed on CIFAR100 and ILSVRC2012. CIFAR100 contains $50\mathrm{K}$ training and $10\mathrm{K}$ testing images. These images have a fixed spatial resolution of $32\times32$, and are uniformly distributed over $100$ classes. ILSVRC2012 is a subset of the ImageNet database~\cite{deng2009imagenet}, which contains $1\rm{,}000$ categories located at different levels of the WordNet hierarchy. The training set has around $1.3\mathrm{M}$ images, which are roughly uniformly distributed over all classes. The testing set ({\em a.k.a.}, the validation set used in the competition) has $50\mathrm{K}$ images, $50$ for each class.

On CIFAR100, we evaluate deep residual networks~\cite{he2016deep} with $20$, $56$ and $110$ layers, respectively. All settings follow the conventions ($164$ epochs, with data augmentation using both random flip and crop), which are detailed in an online repository\footnote{\label{Footnote1}{\sf https://github.com/bearpaw/pytorch-classification}}. The baseline reports classification rates of $67.11\%$, $70.10\%$ and $71.21\%$ for $20$-layer, $56$-layer and $110$-layer networks, and after our approach is applied, these numbers are consistently improved to $\mathbf{69.27}\%$, $\mathbf{71.75}\%$ and $\mathbf{72.31}\%$, respectively.

On ILSVRC2012, we train a deep residual network~\cite{he2016deep} with $18$ layers from scratch. All settings also follow the conventions ($90$ epochs, a series of data augmentation tricks in training, single $224\times224$ crop in testing), which are also detailed online\textsuperscript{\ref{Footnote1}}. With our approach, the top-$1$ accuracy of ResNet-$18$ is boosted from $69.57\%$ to $\mathbf{70.08}\%$.

The accuracy gain in classification seems smaller than object localization, which is partly due to the abundance of training data in classification. In what follows, we show much larger improvements in few-shot experiments.

\renewcommand{\colwidthA}{0.54cm}
\renewcommand{\colwidthB}{1.06cm}
\begin{table}[!btp]
\centering{
\setlength{\tabcolsep}{0.08cm}
\begin{tabular}{|l|L{\colwidthA}||C{\colwidthB}|C{\colwidthB}||C{\colwidthB}|C{\colwidthB}|C{\colwidthB}|}
\hline
{\bf Avg} & TR &         $1$-shot &         $2$-shot &         $5$-shot &        $10$-shot &        $20$-shot \\
\hline\hline
\multirow{2}{*}{{\em novel}} &
      {\bf BL} &          $45.23$ &          $56.90$ &          $68.68$ &          $74.36$ &          $77.69$ \\
\cline{2-7}
{}  & {\bf PT} & $\mathbf{46.27}$ & $\mathbf{57.78}$ & $\mathbf{69.42}$ & $\mathbf{74.86}$ & $\mathbf{78.03}$ \\
\hline
\multirow{2}{*}{{\em all}} &
      {\bf BL} &          $57.65$ &          $64.69$ &          $72.35$ &          $76.18$ &          $78.46$ \\
\cline{2-7}
{}  & {\bf PT} & $\mathbf{58.67}$ & $\mathbf{65.68}$ & $\mathbf{73.08}$ & $\mathbf{76.56}$ & $\mathbf{78.57}$ \\
\hline
\multirow{2}{*}{{\em all+p}} &
      {\bf BL} &          $56.43$ &          $63.41$ &          $70.95$ &          $74.75$ &          $77.00$ \\
\cline{2-7}
{}  & {\bf PT} & $\mathbf{57.63}$ & $\mathbf{64.55}$ & $\mathbf{71.93}$ & $\mathbf{75.51}$ & $\mathbf{77.60}$ \\
\hline\hline
{\bf Att} & TR &         $1$-shot &         $2$-shot &         $5$-shot &        $10$-shot &        $20$-shot \\
\hline\hline
\multirow{2}{*}{{\em novel}} &
      {\bf BL} &          $46.02$ &          $57.51$ &          $69.16$ &          $74.83$ &          $78.11$ \\
\cline{2-7}
{}  & {\bf PT} & $\mathbf{46.37}$ & $\mathbf{57.93}$ & $\mathbf{69.48}$ & $\mathbf{74.98}$ & $\mathbf{78.11}$ \\
\hline
\multirow{2}{*}{{\em all}} &
      {\bf BL} &          $58.16$ &          $65.21$ &          $72.72$ &          $76.50$ &          $78.74$ \\
\cline{2-7}
{}  & {\bf PT} & $\mathbf{58.87}$ & $\mathbf{65.97}$ & $\mathbf{73.34}$ & $\mathbf{76.87}$ & $\mathbf{78.79}$ \\
\hline
\multirow{2}{*}{{\em all+p}} &
      {\bf BL} &          $56.76$ &          $63.80$ &          $71.24$ &          $75.02$ &          $77.25$ \\
\cline{2-7}
{}  & {\bf PT} & $\mathbf{57.89}$ & $\mathbf{64.91}$ & $\mathbf{72.22}$ & $\mathbf{75.79}$ & $\mathbf{77.81}$ \\
\hline
\end{tabular}}\vspace{2mm}
\caption{Few-shot classification results on ImageNet. We evaluate on both average and attention weight generators, and report results on different shots, {\em i.e.}, $1$, $2$, $5$, $10$ and $20$. The baseline ({\bf BL}) is~\cite{gidaris2018dynamic}.}
\label{Tab:ClassificationMiniImageNet}
\end{table}

\subsubsection{Few-shot Classification}
\label{Experiments:Classification:FewShot}

Few-shot classification experiments are performed on Mini-ImageNet~\cite{vinyals2016matching} and ImageNet~\cite{deng2009imagenet}. Mini-ImageNet has $100$ different categories with $600$ images per category, and each image has a size of $84\times84$. For our experiments we use the splits in~\cite{ravi2017optimization} that include $64$ categories for training, $16$ categories for validation, and $20$ categories for testing. On ImageNet, We follow~\cite{hariharan2017low,wang2018low} to split ILSVRC2012 into $389$ base and $611$ novel categories, using $193$ base and $300$ novel categories for cross validation, and the remaining for final evaluation. 

On Mini-ImageNet, we follow the convention to use a deep residual network with $14$ layers, and investigate a popular setting which takes $64$ categories as the base, and performs $5$-way learning using different numbers of training images (shots). We first train a classification model on the base classes with the same learning strategy as in regular experiments, and then fine-tune it on the novel classes either directly ({\bf BL}) or in a progressive manner ({\bf PT}). Note that we do not design specific techniques for few-shot learning, nevertheless, we observe significant accuracy gain over the baseline, boosting the $5$way-$1$shot, $5$way-$5$shot and $5$way-$20$shot rates from $29.59\%$, $50.53\%$ and $64.75\%$ to $\mathbf{38.72}\%$, $\mathbf{54.05}\%$ and $\mathbf{70.93}\%$, respectively. This verifies our motivation, {\em i.e.}, coarse-to-fine learning has the ability of alleviating over-fitting, so that it works better in this few-shot scenario, we obtain much more significant improvements compared to those in regular classification.

On ImageNet, the baseline~\cite{gidaris2018dynamic} trains a recognition model with a $10$-layer residual backbone and a classifier based on cosine similarity in the first training phase. In the second phase, we follow the convention to freeze the backbone and only train the classifier, with access to both novel and base categories. Two different classifiers with average and attention weight generators are evaluated as in the original paper. We use the evaluation metrics in~\cite{wang2018low}, and results are shown in Table~\ref{Tab:ClassificationMiniImageNet}. Again, our approach improves few-shot classification accuracy consistently. Being a generalized training strategy, it achieves comparable improvements to~\cite{gidaris2018dynamic}, a specifically designed few-shot learning algorithm.

\subsection{Medical Image Segmentation}
\label{Experiments:Segmentation}

The third task is medical image segmentation, which serves as an important prerequisite for computer-assisted diagnosis (CAD). We investigate the scenario of CT scans which are easy to acquire yet raise the problem of organ segmentation. We follow~\cite{yu2018recurrent} to use the dataset containing $16$ organs and blood vessels in $200$ abdominal CT scans. We study each organ individually, where $150$ cases are used for training and the remaining $50$ for testing. We measure the segmentation accuracy by computing the Dice-S{\o}rensen coefficient (DSC) for each sample, and report the average and standard deviation over all tested cases.

The baseline model is RSTN~\cite{yu2018recurrent}, a coarse-to-fine approach which deals with each target individually. Here, both $\mathbb{M}^\mathrm{C}$ and $\mathbb{M}^\mathrm{F}$ are fully-convolutional networks (FCN)~\cite{long2015fully}. As mentioned before, we do not need to process the coarse output in this case, so $\mathbf{g}\!\left(\cdot\right)$ contains two convolutional layers on the segmentation mask $\tilde{\mathbf{y}}$, blurring it into a saliency map before adding it to the original image. To filter out less useful input contents, a minimal bounding box is built to cover all pixels with a probability of at least $0.5$, and the input image is cropped accordingly before fed into the fine stage. Different from the previous baselines, RSTN works in an iterative manner, {\em i.e.}, more than one fine stages are performed during testing, until convergence is achieved.

In the original paper~\cite{yu2018recurrent}, to improve training stability, the authors designed a stepwise training strategy which first feeds the ground-truth mask $\mathbf{y}^\star$ into $\mathbb{M}^\mathrm{F}$, and changes it to $\mathbf{y}^\mathrm{C}$ at a fixed point of the training process. However, it still failed to converge on $3$ out of $16$ targets (see Table~\ref{Tab:Segmentation}), and had a probability to fail on other $5$. By applying progressive training, we allow the supervision signal to change gradually from $\mathbf{y}^\star$ to $\mathbf{y}^\mathrm{C}$, not suddenly. There are in total $120\mathrm{K}$ iterations with a mini-batch size of $1$. Because semantic segmentation is much more difficult than previous tasks, we use ${t}\equiv{0}$ in the first $40\mathrm{K}$ iterations otherwise coarse prediction may provide a meaningless mask and thus totally confuse the fine stage. We change $t$ gradually from $0$ to $1$ in the next $40\mathrm{K}$ iterations, and set ${t}\equiv{1}$ in the last $40\mathrm{K}$ iterations. The learning rate starts with $10^{-5}$ and is divided by $2$ after $90\mathrm{K}$, $100\mathrm{K}$ and $110\mathrm{K}$ iterations.

\renewcommand{\colwidthA}{1.1cm}
\begin{table}[!btp]
\centering{
\setlength{\tabcolsep}{0.12cm}
\begin{tabular}{|l||R{\colwidthA}|R{\colwidthA}||R{\colwidthA}|R{\colwidthA}|}
\hline
Organ                        & {\bf BL}-{\bf C} & {\bf BL}-{\bf F} & {\bf PT}-{\bf C} & {\bf PT}-{\bf F} \\
\hline\hline
{\em aorta}$^{\ast}$         &          $90.78$ &   $^\sharp90.76$ &          $91.99$ & $\mathbf{93.69}$ \\
\hline
{\em adrenal gland}          &          $60.70$ &          $63.76$ &          $60.97$ & $\mathbf{64.11}$ \\
\hline
{\em celiac a.a.}$^{\ast}$   &              $-$ &              $-$ &          $52.96$ & $\mathbf{56.02}$ \\
\hline
{\em colon}                  &          $74.69$ &   $^\sharp74.14$ &          $77.72$ & $\mathbf{79.94}$ \\
\hline
{\em duodenum}               &          $71.40$ & $\mathbf{73.42}$ &          $67.80$ &          $72.22$ \\
\hline
{\em gallbladder}            &          $87.08$ &          $87.10$ &          $87.58$ & $\mathbf{89.68}$ \\
\hline
{\em inferior v.c.}$^{\ast}$ &          $79.12$ &          $79.69$ &          $81.54$ & $\mathbf{83.39}$ \\
\hline
{\em kidney left}            &          $96.08$ &          $96.21$ &          $95.64$ & $\mathbf{96.21}$ \\
\hline
{\em kidney right}           &          $95.80$ & $\mathbf{95.97}$ &          $95.35$ &          $95.79$ \\
\hline
{\em liver}                  &          $96.70$ &          $96.75$ &          $96.28$ & $\mathbf{96.93}$ \\
\hline
{\em pancreas}               &          $86.09$ &          $87.60$ &          $83.66$ & $\mathbf{87.91}$ \\
\hline
{\em superior m.a.}$^{\ast}$ &              $-$ &              $-$ &          $66.89$ & $\mathbf{74.01}$ \\
\hline
{\em small bowel}            &          $63.86$ &   $^\sharp63.52$ &          $71.90$ & $\mathbf{75.32}$ \\
\hline
{\em spleen}                 &          $96.61$ &          $96.78$ &          $96.02$ & $\mathbf{96.80}$ \\
\hline
{\em stomach}                &          $94.82$ & $\mathbf{94.98}$ &          $93.53$ &          $94.59$ \\
\hline
{\em veins}$^{\ast}$         &              $-$ &              $-$ &          $72.89$ & $\mathbf{75.13}$ \\
\hline\hline
{\bf average}: {\em organs}  &          $83.98$ &          $84.57$ &          $84.22$ & $\mathbf{86.32}$ \\
\hline
{\bf average}: {\em vessels} &              $-$ &              $-$ &          $73.26$ & $\mathbf{76.45}$ \\
\hline
{\bf average}: {\em all}     &              $-$ &              $-$ &          $80.80$ & $\mathbf{83.23}$ \\
\hline
\end{tabular}}\vspace{2mm}
\caption{
Comparison of coarse ({\bf C}) and fine ({\bf F}) segmentation by the baseline ({\bf BL}, \cite{yu2018recurrent}) and the proposed progressive training strategy ({\bf PT}). A target is marked by an asterisk if it is a blood vessel. The original version of RSTN~\cite{yu2018recurrent} cannot achieve convergence on three blood vessels (marked by $-$). A fine-scaled accuracy is indicated by $\sharp$ if it is lower than the coarse-scaled one.
}
\label{Tab:Segmentation}
\end{table}

Results are summarized in Table~\ref{Tab:Segmentation}. We can see that, after our approach is applied, RSTN achieves convergence on all $16$ targets, including the $3$ blood vessels which failed to converge. Among all $16$ targets, our approach improves the segmentation accuracy of $9$ of them, with $5$ of them over $1\%$, $3$ of them over $3\%$ and $1$ of them over $10\%$ ({\em small bowel}). Slight accuracy drop ($<0.40\%$) is reported on $2$ out of $16$ targets, and the maximal drop is $1.20\%$ ({\em duodenum}). In general, the average accuracy over $13$ converged targets is boosted by $2\%$, which is significant given such a high baseline and the fact that our approach merely changes the original training strategy of RSTN.

{\bf The success on RSTN shows that our approach can be freely applied to iterative learning frameworks.}

\section{Conclusions}
\label{Conclusions}

In this paper, we present the first coarse-to-fine (C2F) learning framework that applies to a wide range of vision problems. We make two contributions, namely, encoding the prediction into a dense matrix to concatenate with the original input, and designing a progressive training strategy to achieve both stability in training and higher accuracy. The effectiveness of our approach is verified on three popular vision tasks, {\em i.e.}, classification, localization and segmentation. In particular, the ability of our approach at alleviating over-fitting is shown in training curves, and highlighted in few-shot experiments.

This paper leaves much room for future research. For example, the strategy of monotonically increasing the difficulty of training data may not be perfect, as some prior work~\cite{huang2018snapshot} verified that disturbing the training process can lead to better model ensemble. This is possibly related to some specific strategies for each particular task. Studying these topics may provide new perspectives to understand machine learning, in particular deep learning methods.

{\small
\bibliographystyle{ieee}
\bibliography{egpaper_final}
}

\end{document}